\documentclass[letterpaper, 10 pt, conference]{ieeeconf}

\IEEEoverridecommandlockouts                              

\usepackage{graphicx} 
\graphicspath{{images/}{generated-images/}{results/}}
\DeclareGraphicsExtensions{.pdf,.jpeg,.png}

\usepackage{amsmath} 
\usepackage{bm}
\usepackage{amssymb}  

\usepackage[caption=false,font=footnotesize]{subfig}

\usepackage{url}
\usepackage{array}
\usepackage{hyperref}
\usepackage{cite}
\usepackage{color}

\newcommand{\M}[1]{{\bm #1}} 
\newcommand{\robots}{ \M{\Omega}}
\newcommand{\vxx}{\boldsymbol{x}}
\newcommand{\of}[2]{_{#1_{#2}}}
\newcommand{\SEthree}{\ensuremath{\mathrm{SE}(3)}}
\newcommand{\MR}{\M{R}}

\newcommand{\vt}{\boldsymbol{t}}
\newcommand{\SOthree}{\ensuremath{\mathrm{SO}(3)}}
\newcommand{\Real}[1]{ { {\mathbb R}^{#1} } }

\newcommand{\vs}{\boldsymbol{s}}
\renewcommand{\vs}{{\tt{vs}}}
\newcommand{\beq}{\begin{equation}}
\newcommand{\eeq}{\end{equation}}

\newcommand{\link}[4]{^{#1_{#2}}_{#3_{#4}}}
\newcommand{\subs}{\link{\alpha}{i}{\beta}{j}}
\newcommand{\MRbar}{\bar{\MR}}

\newcommand{\vz}{\boldsymbol{z}}
\newcommand{\vzbar}{\bar{\vz}}

\newcommand{\vtbar}{\bar{\vt}}
\newcommand{\tran}{^{\mathsf{T}}}

\newcommand{\omegat}{{\tiny \omega_t^2}}
\newcommand{\omegaR}{{\tiny \omega_R^2}}
\newcommand{\eye}{{\mathbf I}}
\newcommand{\bea}{\begin{eqnarray}}
\newcommand{\eea}{\end{eqnarray}}
\newcommand{\calE}{{\cal E}}
\newcommand{\normsq}[2]{\left\|#1\right\|^2_{#2}}
\newcommand{\frob}{{\tt F}}
\newcommand{\vr}{\boldsymbol{r}}
\newcommand{\vp}{\boldsymbol{p}}
\newcommand{\vy}{\boldsymbol{y}}
\newcommand{\vg}{\boldsymbol{g}}
\newcommand{\MH}{\M{H}}
\newcommand{\MOmega}{\M{\Omega}}
\newcommand{\at}[1]{^{({#1})}}
\newcommand{\inv}{^{-1}}

\newcommand\norm[1]{\left\lVert#1\right\rVert}
\newcommand\ltnorm[1]{\norm{#1}_{\ell_2}}

\newcommand{\calF}{{\cal F}}
\newcommand{\vv}{\boldsymbol{v}}
\newcommand{\vc}{\boldsymbol{c}}

\newcommand{\reffig}[1]{Fig.~\ref{#1}}

\newcommand{\reftab}[1]{Table~\ref{#1}}
\newcommand{\refsec}[1]{Section~\ref{#1}}

\definecolor{todo-red}{RGB}{200,12,12}
\definecolor{green4}{RGB}{0,128,0}

\DeclareMathOperator*{\argmin}{arg\,min}

\title{\LARGE \bf Data-Efficient Decentralized Visual SLAM}

\author{Titus Cieslewski$^{1}$, Siddharth Choudhary$^{2}$ and Davide Scaramuzza$^{1}$%
\thanks{$^{1}$T. Cieslewski and D. Scaramuzza are with the Robotics and Perception Group, Dep. of Informatics, University of Zurich , and Dep. of Neuroinformatics, University of Zurich and ETH Zurich, Switzerland---\url{http://rpg.ifi.uzh.ch.}}%
\thanks{$^{2}$S. Choudhary is with the College of Computing, Georgia Institute of Technology, Atlanta, GA, USA, \url{siddharth.choudhary@gatech.edu}}%
}

\begin{document}

\maketitle

\begin{abstract}

Decentralized visual simultaneous localization and mapping (SLAM) is a powerful tool for multi-robot applications in environments where absolute positioning systems are not available.
Being visual, it relies on cameras, cheap, lightweight and versatile sensors, and being decentralized, it does not rely on communication to a central ground station.
In this work, we integrate state-of-the-art decentralized SLAM components into a new, complete decentralized visual SLAM system.
To allow for data association and co-optimization, existing decentralized visual SLAM systems regularly exchange the full map data between all robots, incurring large data transfers at a complexity that scales quadratically with the robot count.
In contrast, our method performs efficient data association in two stages:
in the first stage a compact full-image descriptor is deterministically sent to only one robot.
In the second stage, which is only executed if the first stage succeeded, the data required for relative pose estimation is sent, again to only one robot.
Thus, data association scales linearly with the robot count and uses highly compact place representations.
For optimization, a state-of-the-art decentralized pose-graph optimization method is used.
It exchanges a minimum amount of data which is linear with trajectory overlap.
We characterize the resulting system and identify bottlenecks in its components.
The system is evaluated on publically available data and we provide open access to the code.

\end{abstract}

\section{Introduction}

Using several robots instead of one can accelerate many tasks such as exploration and mapping, or enable heterogenous teams of robots, where each robot has a specialization.
Multi-robot simultaneous localization and mapping (SLAM) is an essential component of any team of robots operating in an absolute positioning system denied environment, as it relates the current state estimate of each robot to all present and past state estimates of all robots.
Because cameras are cheap, light-weight and versatile sensors, we seek to implement a visual SLAM system.
Visual Multi-Robot SLAM can be solved in a centralized manner, where a single entity collects all data and solves SLAM for all robots, but that relies on a central entity to always be reachable, to never fail and to scale to the size of the robot team, both in computation and bandwidth.
Decentralized systems do not have this bottleneck, but are more challenging to implement.
%
%
%
%
%
%
%
%

\begin{figure}
  \centering
  \subfloat[Ten sub-trajectories of KITTI 00 after running our method.
  Each color represents an individual trajectory, place matches are marked in black, dashed lines indicate the ground truth.\label{kitti-60}]{
    \includegraphics[width=.8\columnwidth]{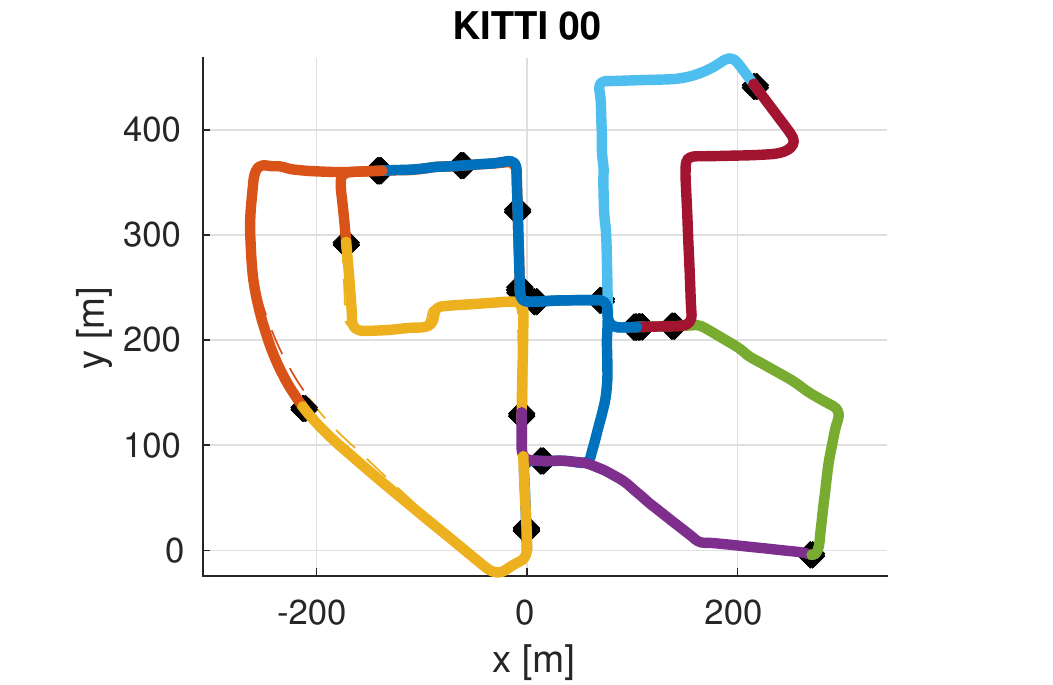}}
    
  \subfloat[Data transmission over time for the system components:
  decentralized optimization (DOpt), decentralized visual place recognition (DVPR) and relative pose estimation (RelPose).\label{data-time-60}]{
    \includegraphics[width=.8\columnwidth]{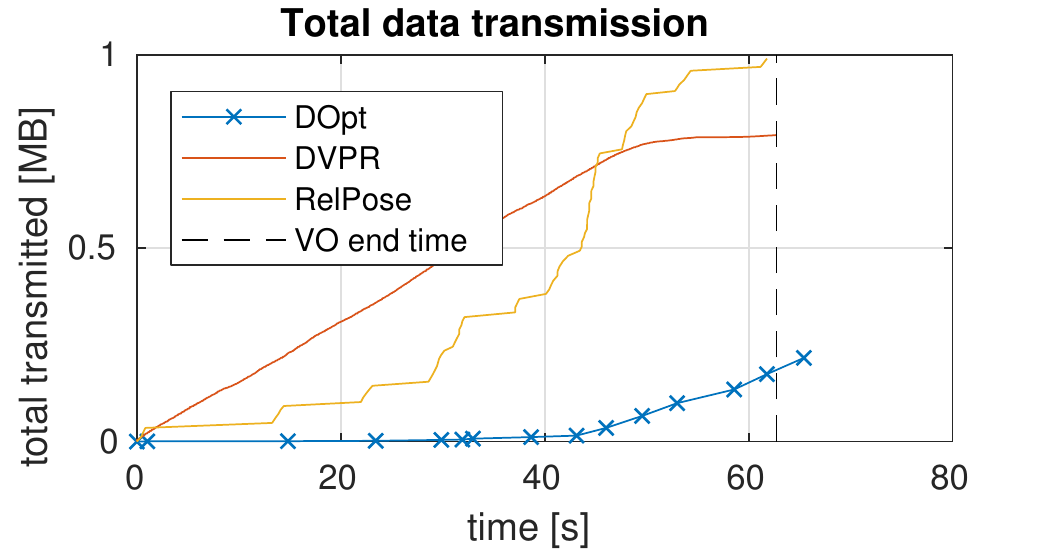}}
  \caption{The proposed decentralized visual SLAM system is able to build a globally consistent map from ten sub-trajectories of KITTI 00 by exchanging only around 2MB in total -- for decentralized optimization, place recognition and visual relative pose estimation, for all ten robots.}
  \vspace{-5mm}
\end{figure}

%
%
%
%
%
%
%

Visual SLAM systems typically consist of three components:
1) a visual odometry algorithm which provides an initial state estimate,
2) a place recognition system which is able to relate the currently observed scene to previously observed scenes, and 
3) an optimization back-end which consistently integrates the place matches from the place recognition system with the full stream of  state estimates.
The end product is a map, and that map feeds back to place recognition and optimization.
It is this feedback which makes decentralized SLAM challenging, especially if one is concerned about communication bandwidth.
Visual odometry is not involved in the feedback loop and is thus trivial to distribute -- it can be run on each robot independently and feed its output to the rest of the decentralized SLAM system.

In previous work we have proposed state-of-the art decentralized place recognition \cite{cieslewski2017efficient2} and optimization \cite{choudhary2016distributed} systems separately.
Both systems focus on data efficiency: they achieve performance similar to a centralized system while minimizing the data that needs to be exchanged.
In this work, we integrate both systems along with a data-efficient method for visual feature association \cite{tardioli2015visual} into a full decentralized visual SLAM system.
The system is evaluated on publically available data and the code is provided open-source.

\section{Related Work}

In the following, we independently consider related work pertaining to decentralized optimization and to decentralized place recognition.
We conclude with a review of existing integrated decentralized SLAM systems.

\subsection{Decentralized Optimization}\label{optrw}

Distributed estimation in multi-robot systems is an active field of research,
with special attention being paid to communication constraints~\cite{Paull15icra},
heterogeneity~\cite{Bailey11icra,Indelman12ijrr}, consistency~\cite{Bahr09icra},
 and robust data association~\cite{Dong15icra}.
The literature offers distributed implementations of different estimation techniques,
including Kalman filters~\cite{Roumeliotis02tra},
information filters~\cite{Thrun03isrr}, 
particle filters~\cite{Howard06ijrr,Carlone11jirs}, and distributed smoothers~\cite{Cunningham13icra, choudhary2016distributed}
%

In multi-robot systems, maximum-likelihood trajectory estimation can be
performed by collecting all measurements at a centralized inference engine, which
performs the optimization~\cite{Andersson08icra,Kim10icra,Bailey11icra,Lazaro13iros,Dong15icra}.
However, it is not practical to collect all measurements at a single inference engine since it requires a large communication bandwidth. 
Furthermore, solving trajectory estimation over a large team of robots can be too demanding for a
single computational unit.

These reasons triggered interest towards \emph{distributed trajectory estimation}, in which the robots
only exploit local communication, in order to reach a consensus on the trajectory estimate~\cite{Nerurkar09icra,Franceschelli10icra,Aragues11icra,Knuth13icra,Cunningham10iros}.
Recently, Cunnigham et al.~\cite{Cunningham10iros,Cunningham13icra} used Gaussian elimination, and developed an approach,
called DDF-SAM, in which robots exchange Gaussian marginals over the~\emph{separators} (i.e., the variables observed by
multiple robots).
%

While Gaussian elimination has become a popular approach,
it has two major shortcomings. First, the marginals to be exchanged among the robots are dense, hence
the communication cost is quadratic in the number of separators.
This motivated the use of sparsification techniques~\cite{Paull15icra}.
Second, Gaussian elimination is performed on a linearized version of the
 problem, hence these approaches require good linearization points and
complex bookkeeping to ensure consistent linearization across the robots~\cite{Cunningham13icra}.
The need of a linearization point also characterizes gradient-based techniques~\cite{Knuth13icra}.
An alternative approach to Gaussian elimination is the Gauss-Seidel approach of Choudhary et al.~\cite{choudhary2016distributed},
which requires communication linear in the number of separators. The Distributed Gauss-Seidel approach only requires the latest estimates for 
poses involved in inter-robot measurement to be communicated and therefore does not require consistent linearization and complex bookkeeping.
This approach also avoids information double counting since it does not communicate marginals.

%

\subsection{Decentralized Place Recognition and Pose Estimation}

\begin{figure}
  \centering
  \includegraphics[width=.5\columnwidth]{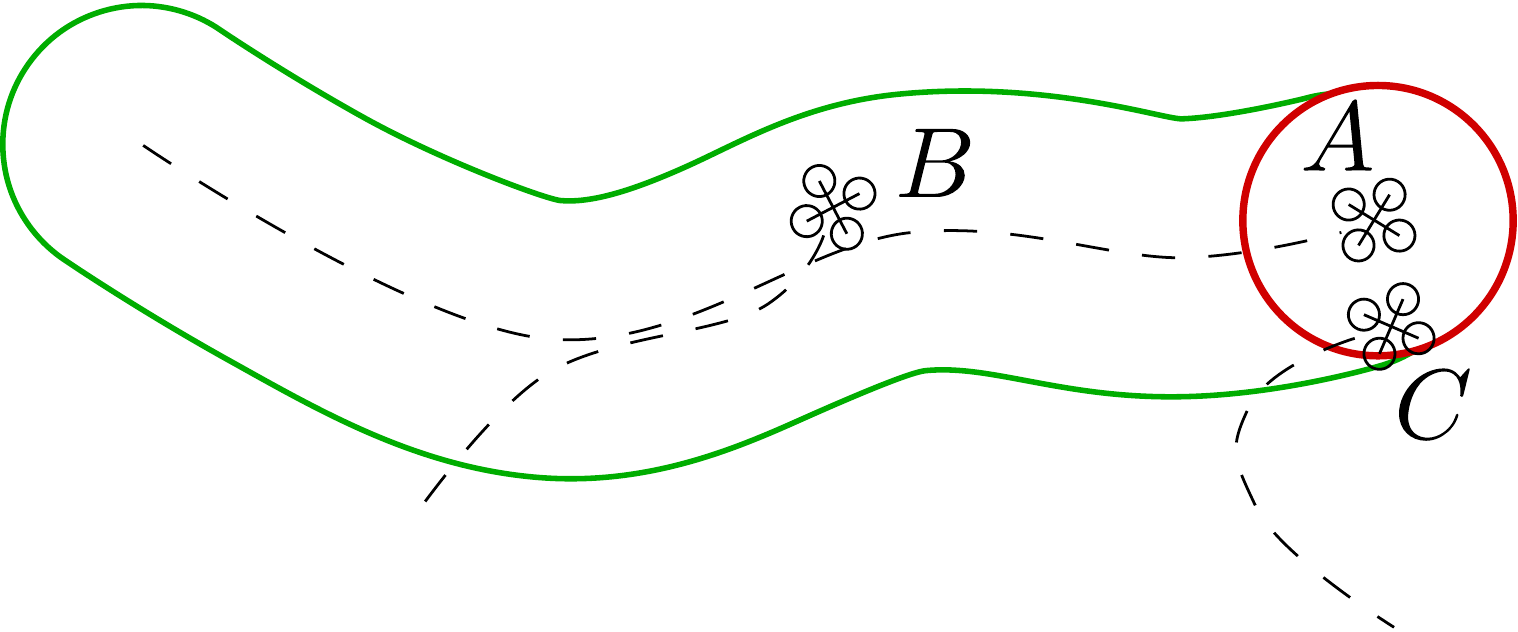}
  \vspace{-3mm}
  \caption{Restricting relative pose measurements to robots that are in line of sight (red circle) severely limits the recall of relative localization:
  Robots $A$ and $C$ can establish relative localization but $A$ and $B$ cannot.
  This often applies to direct relative measurements.
  By ensuring that relative localization can occur from all past measurements (green outline) we can increase recall and put less restrictions on the high-level application that controls the robots.}
  \label{fig:loslimit}
  \vspace{-4mm}
\end{figure}

In order to solve decentralized SLAM, measurements that relate poses between different robots ({\it inter-robot measurements}) need to be established.
A popular type of inter-robot measurements are direct measurements of the other robot \cite{Zhou06iros}, such as time-of-flight distance measurements \cite{Paull15icra} or vision-based relative pose etimation \cite{Kim10icra}.
The latter is typically aided with markers that are deployed on the robots.
To the best of our knowledge, most types of direct measurements require specialized hardware (which can precisely measure time-of-flight for example, or visual markers).
Furthermore, many types of direct measurements require the robots to be in line of sight, which, in many environments, imposes a major limitation on the set of relative poses that can be established, see \reffig{fig:loslimit}.
Limiting relative measurements in this way translates into limitations for higher-level applications that would like to use decentralized SLAM as a tool.

We prefer, instead, to use indirect relative measurements.
These are established through registration of observations that two robots $A$ and $B$ make of the same scene \cite{aragues2011consistent, tardioli2015visual}.
This requires no additional hardware:
the sensors already used for odometry can be re-used for inter-robot measurements.
Since establishing inter-robot measurements in this way is equivalent to establishing loop closures in single-robot SLAM, indirect measurements are widely used in centralized visual SLAM algorithms \cite{forster2013collaborative, riazuelo2014c, schmuck2017multi}.
The establishment of indirect measurements is still limited by the communication range of the robots, but in practice it is feasible to establish a communication range that exceeds line-of-sight.
Furthermore, if two robots cannot communicate directly, they might still communicate through a multi-hop network.
A disadvantage of indirect measurements, then, is that they require bandwidth to communicate.
If all robots share all data with every other robot, the amount of data exchanged is quadratic in the number of robots.
One way to mitigate the bandwidth is to exchange data only with immediate neighbours, but this has the same reduced recall as discussed in \reffig{fig:loslimit}.
Another way to reduce bandwidth is to use compact representations that allow to establish inter-robot measurements with a minimal amount of data exchange.
Visual feature-based maps can be compressed by compressing feature descriptors \cite{lynen2015get}, substituting feature descriptors for corresponding cluster centers in a {\it visual vocabulary} \cite{tardioli2015visual}, removing landmarks that prove not to be necessary for localization \cite{dymczyk2015gist, contreras2017O} or using compact full-image descriptors \cite{arandjelovic2016netvlad}.
In previous work, we have shown another way to reduce the bandwidth of place recognition:
data exchange can be reduced by a factor of the robot count if the problem can be cast to key-value lookup\cite{cieslewski2017efficient}.
In this work, we use \cite{cieslewski2017efficient2}, an advanced version of \cite{cieslewski2017efficient} which at the same time makes use of a compact, state-of-the-art place representation, NetVLAD \cite{arandjelovic2016netvlad}.
\cite{tardioli2015visual} is used in association of visual features for relative pose estimation.

\subsection{Integrated Decentralized SLAM}

Several systems have previously combined decentralized optimization and relative pose estimation into decentralized SLAM.
Several optimization works cited in \refsec{optrw}, such as \cite{Paull15icra, choudhary2016distributed} use direct relative measurements and can thus be considered full decentralized SLAM systems with the recall limitation illustrated in \reffig{fig:loslimit}.
DDF-SAM \cite{Cunningham10iros} has been extended to a decentralized SLAM system in \cite{Cunningham12icra} by adding a data association system which matches planar landmark points.
This system has been validated in the real world, but the landmarks consisted of manually placed poles that were detected by a laser range finder.
\cite{Lazaro13iros} establishes relative poses by exchanging and aligning 2D laser scan edges.
\cite{choudhary2016distributed} has been extended to a decentralized SLAM system in \cite{choudhary2017distributed} where data association is provided by identification and pose estimation of commonly observed objects.
This approach relies on the existence of unique, pre-trained objects in the environment.
The majority of decentralized SLAM systems we have found in our literature review are not vision-based.
A first decentralized visual SLAM system has been proposed in \cite{bresson2013consistent}, but it relies on exchanging the full maps between robots.
Furthermore, it is only evaluated on $50m \times 50m$ L-shaped trajectory and with only two robots, with images obtained from a simulated environment.
In contrast, we evaluate our system on a large scale scenario, with more robots, on real data, and our system uses state-of-the-art algorithms which exchange far less than the full maps.

\section{Methodology}
\label{sec:methodology}

\begin{figure}
  \centering
  \includegraphics[width=.8\columnwidth]{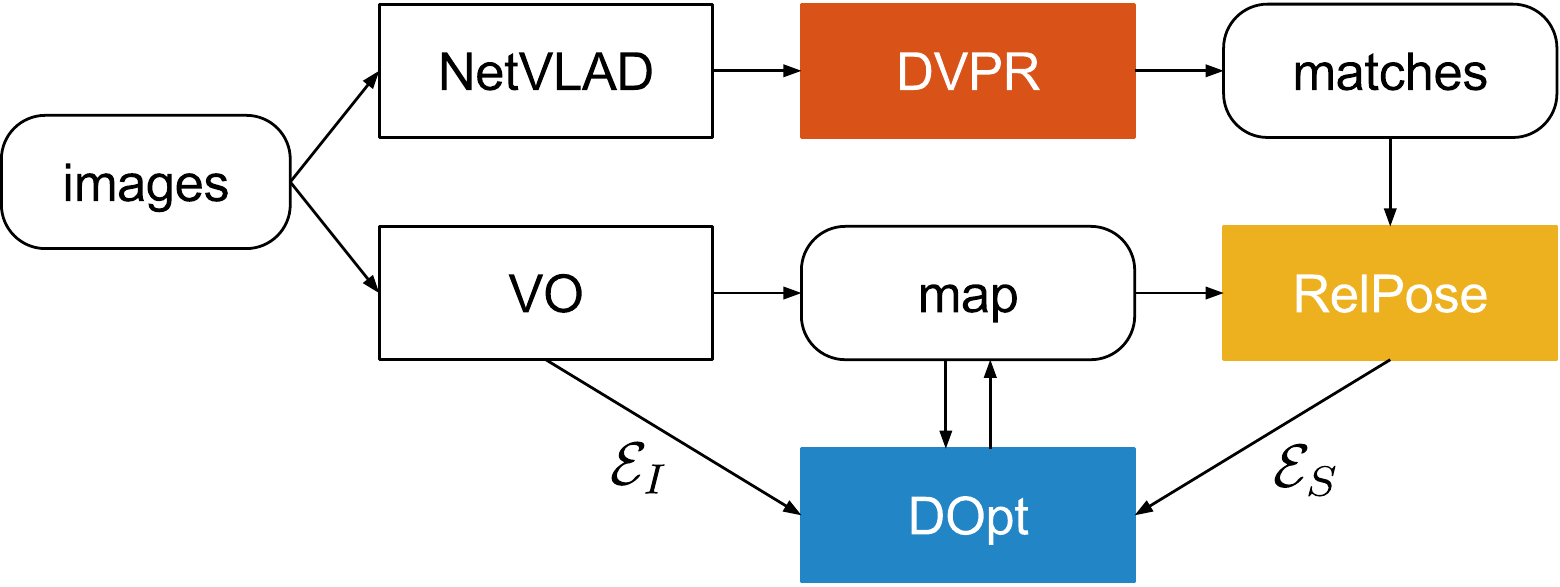}
  \caption{Components and interactions in our system.
The diagram shows the components that run on each robot.
The colored blocks indicate components that communicate with other robots.
Sharp corners indicate software modules, rounded corners indicate data.
{\it DVPR}: Decentralized visual place recognition, {\it VO}: visual odometry, {\it RelPose}: Geometric verification and relative pose estimation, {\it DOpt}: Decentralized optimization.}
  \label{fig:system}
  \vspace{-4mm}
\end{figure}
The proposed decentralized SLAM system consists of $|\MOmega|$ robots $\MOmega =\{\alpha, \beta, \gamma, \ldots\}$ which each execute the system illustrated in \reffig{fig:system}.
The images taken by a camera are fed to both NetVLAD \cite{arandjelovic2016netvlad} and a visual odometry (VO).
NetVLAD produces a compact image representation used for decentralized visual place recognition (DVPR) \cite{cieslewski2017efficient2}, which itself produces candidate place recognition matches.
The VO produces a sparse feature map that can be used for localization.
The relative pose estimation module (RelPose) extracts the parts of the sparse map that are observed at the candidate place matches and uses them to establish relative poses between the robots trajectories, or to reject candidate matches.
The decentralized optimization module (DOpt) \cite{choudhary2016distributed} obtains the initial guess from the map, intra-robot relative pose measurements $\mathcal{E}_I$ from the VO and the inter-robot relative pose measurements $\mathcal{E}_S$ from RelPose, and updates the map.
The system works continuously as new images are acquired.

\subsection{Intra-Robot Measurements}\label{sec:vo}

Intra-robot measurements are obtained from a visual odometry (VO) algorithm.
Any VO algorithm can be used in our system as long as it fulfills the following requirements:

{\it 1) It produces a pose graph.}
The VO of robot $\alpha$ defines frames $\calF_\alpha = \{(\alpha, 1), (\alpha, 2), (\alpha, 3), ...\}$ and provides the intra-robot measurements
\begin{align}
\calE_{I, \alpha} &\doteq \{ \vzbar\link{\alpha}{i}{\alpha}{i+1} \forall (\alpha, i), (\alpha, i+1) \in \calF_\alpha \} \\
\vzbar\link{\alpha}{i}{\alpha}{i+1} &\doteq (\MRbar\link{\alpha}{i}{\alpha}{i+1}, \vtbar\link{\alpha}{i}{\alpha}{i+1}) \in \SEthree \label{eq:odom}
\end{align}
describing pose transforms between subsequent frames.

{\it 2) Each pose is associated to an image to be used for place recognition}.
We denote this image by $I\of{\alpha}{i}$.
Our current system does not yet take full advantage of a multi-camera setup, which could be a useful future extension.
The image capture time is denoted as $t\of{\alpha}{i}$
\footnote{A common time reference is hard to establish, and, strictly speaking, not well defined, so consider $t\of{\alpha}{i}$ a Lamport timestamp \cite{lamport1978time}.
$t\of{\alpha}{i}$ is not actually used in the implementation, so this detail is not particularly important.}.

{\it 3) The VO provides absolute scale,} as implied by \eqref{eq:odom}.
Monocular VO typically estimates the pose up to a scale due to scale ambiguity.
%
%
%
Scale can be obtained from a stereo camera or by fusing with inertial measurements.

In our implementation, we use for visual odometry ORB-SLAM \cite{mur2015orb} in stereo configuration.

\subsection{Inter-Robot Measurements}

The inter-robot measurements are
\beq
\calE_S \doteq \{ \vzbar\subs | \alpha \neq \beta \}
\eeq
with $\vzbar\subs$ defined like \eqref{eq:odom}.
They are established in two phases:
Given its latest image $I\of{\alpha}{i}$, robot $\alpha$ first tries to determine whether there is any image $I\of{\beta}{j}$ of the same scene previously captured by another robot.
This is achieved using decentralized visual place recognition, detailed in \refsec{sec:dvpr}.
Then, if $\exists I\of{\beta}{j}$, $\alpha$ tries to establish $\vzbar\subs$ using the method detailed in \refsec{sec:relpose}.
This two-phase approach is pursued because, as we will see, establishing relative poses consumes more data than recognizing whether $\exists I\of{\beta}{j}$.
The first phase is capable of reducing the amount of data to be exchanged by determining which robot to send the relative pose query to, and by determining whether it is worthwhile to send it in the first place, using a minimal amount of data.

\subsection{Decentralized Visual Place Recognition}\label{sec:dvpr}

The first phase is accomplished by matching image $I\of{\alpha}{i}$ to the images previously captured by the other robots, $\{ I\of{\gamma}{k} | t\of{\gamma}{k} < t\of{\alpha}{i}\}$, using the full-image descriptor NetVLAD \cite{arandjelovic2016netvlad}.
NetVLAD uses a neural network to calculate a vector $\vv\of{\alpha}{i} \in \mathbb{R}^{D_\text{NetVLAD}}$ from $I\of{\alpha}{i}$.
In rough terms, it is trained such that $\ltnorm{\vv\of{\alpha}{i} - \vv\of{\gamma}{k}}$ is lower if $I\of{\alpha}{i}$ and $I\of{\gamma}{k}$ observe the same scene than if they observe different scenes.
For more details, see \cite{arandjelovic2016netvlad}.
NetVLAD is well-suited for decentralized place recognition, as $\vv\of{\alpha}{i}$ is all that robot $\alpha$ needs to send to robot $\gamma$ to establish whether a matching image exists among $\{I\of{\gamma}{k}\}$.
Concretely, we seek the image $I\of{\beta}{j}^\star$ whose NetVLAD vector $\vv\of{\beta}{j}^\star$ has the shortest $\ell_2$ distance to $\vv\of{\alpha}{i}$ and which satisfies
\beq
\ltnorm{\vv\of{\alpha}{i} - \vv\of{\beta}{j}} < \tau_\text{NetVLAD} \label{eq:nvthresh}
\eeq
where $\tau_\text{NetVLAD}$ is a threshold parameter.
$I\of{\beta}{j}^\star$ can be found by sending $\vv\of{\alpha}{i}$ to all other robots, but we use a method that requires $|\MOmega|$ times less data exchange \cite{cieslewski2017efficient2}, which is particularly interesting for scaling to large robot teams.
In this method, each robot $\gamma$ is pre-assigned a cluster center $\vc_\gamma$ and each robot knows the $\vc$ of all other robots.
The query $\vv\of{\alpha}{i}$ is sent only to robot $\delta = \argmin_{\gamma} \ltnorm{\vv\of{\alpha}{i} - \vc_\gamma}$, and $\delta$ replies with the identifier of $I\of{\beta}{j}$,
\begin{equation}
(\beta, j) = \argmin_{(\gamma, k) : \vv\of{\gamma}{k} \in \mathcal{V}_{\delta}} \ltnorm{\vv\of{\alpha}{i} - \vv\of{\gamma}{k}},
\end{equation}
if $\vv\of{\beta}{j}$ also satisfies \eqref{eq:nvthresh}, where $\mathcal{V}_{\delta}$ is the Voronoi cell
$\mathcal{V}_{\delta} = \{\vv | \vc_\delta = \argmin_{\vc_\gamma} \ltnorm{\vv - \vc_\gamma} \}$.
The query is executed for every frame of every robot, so at $t\of{\alpha}{i}$ this query has already been executed $\forall \gamma, k : t\of{\gamma}{k} < t\of{\alpha}{i}$, and $\delta$ is aware of all $\vv\of{\gamma}{k} \in \mathcal{V}_{\delta}$ by virtue of having received the previous queries.
It can thus provide $(\beta, j)$ without any extra data exchange.
This method approximates $I\of{\beta}{j}^\star \sim I\of{\beta}{j}$;
$I\of{\beta}{j}^\star = I\of{\beta}{j}$ if $\exists \delta : \vv\of{\beta}{j}^\star, \vv\of{\alpha}{i} \in \mathcal{V}_\delta$.
In the method, the cluster centers are determined using k-means clustering on a training dataset, in our case the Oxford Robotcar Dataset \cite{maddern20171}.
The properties of the method are discussed in more detail in \cite{cieslewski2017efficient2}.

\subsection{Relative Pose Estimation}\label{sec:relpose}

Once and if $(\beta, j)$ has been established, $\alpha$ sends a relative pose estimation request to $\beta$.
The relative pose estimation (RelPose) can and should heavily depend on the visual odometry (VO) since it can benefit from re-using by-products of the VO such as 3D positions of landmarks.
In our method, we use ORB-SLAM \cite{mur2015orb} for VO and accordingly, our RelPose implementation imitates its loop closure method.
However, it does not imitate \cite{mur2015orb} exactly, since the latter makes use of the data surrounding the match, which in decentralized RelPose would translate into a lot of data exchange.
We want to avoid this and thus, we combine the relative pose estimation method form \cite{mur2015orb} with the visual data association method from \cite{tardioli2015visual}, which reduces the data to be sent from $\alpha$ to $\beta$, $d^{\alpha \rightarrow \beta}_\text{RelPose}$, to a set of landmarks and visual word identifiers:
\begin{equation}
d^{\alpha \rightarrow \beta}_\text{RelPose} = (\{(w_k, \vp_k) \forall k \in K\of{\alpha}{i}\}, \alpha, i, j)
\label{exchdata}
\end{equation}
where $K\of{\alpha}{i}$ are the keypoints observed in $I\of{\alpha}{i}$, $w_k$ is the identifier of a visual word associated to keypoint $k$ and $\vp_k$ is the 3D position of the landmark corresponding to keypoint $k$, expressed in the camera frame of $I\of{\alpha}{i}$.
Visual words represent Voronoi cells in the keypoint descriptor space, here the space of ORB descriptors \cite{rublee2011orb}.
They provide a quantization of the high-dimensional descriptor space, and since similar descriptors are likely to be assigned to the same word, word identifiers can be used for matching keypoints between images \cite{tardioli2015visual}.
Since this is less precise than using raw descriptors, we provide an option to exchange full descriptors instead.
Given $d^{\alpha \rightarrow \beta}$, $\beta$ establishes pairs of matching keypoints $\{(k, k') | w_k = w_{k'}, \nexists ~ l \in K\of{\alpha}{i}: w_l = w_k, \nexists ~ l' \in K\of{\beta}{j}: w_{l'} = w_{k'} \}$.
Given the corresponding pairs of landmark positions $\{(\vp_{k}, \vp_{k'})\}$, RANSAC is used to determine an initial estimate of $\vzbar\subs$ and the corresponding inlier matches $M = \{(k, k')^\star\}$.
A place match is rejected if $|M| < \tau_\text{inliers}$.
Otherwise, the inlier landmark position pairs $\{(\vp_{k}, \vp_{k'})^\star\}$ are used to refine $\vzbar\subs$ to the pose that minimizes a robust 3D registration error
\begin{equation}
\sum_{\{(\vp_{k}, \vp_{k'})^\star\}} \rho_{\tau_\text{loss}} (\ltnorm{\vp_{k} - (\MRbar \cdot \vp_{k'} + \vtbar)}^2).
\end{equation}
We use a robust weight loss function $\rho_{\tau_\text{loss}}(s) = \arctan \frac{s}{\tau_\text{loss}}$ to reduce the weight of remaining keypoint match outliers.

To avoid relative pose $\vzbar\subs$ outliers we use a consistency check, where $\vzbar\subs$ is only accepted if either:
{\it a)} there is another already accepted $\vzbar\link{\alpha}{i'}{\beta}{j'}$ such that $(\alpha, i')$ and $(\alpha, i)$ are within a distance of $\tau_\text{cdist}$ and $\vzbar\subs$ and $\vzbar\link{\alpha}{i'}{\beta}{j'}$ are consistent;
{\it b)} no accepted $\vzbar\link{\alpha}{i'}{\beta}{j'}$ exists, but a previous candidate which fulfills the same conditions.
$\vzbar\subs$ and $\vzbar\link{\alpha}{i'}{\beta}{j'}$ are considered consistent if the two ways of calculating the relative position between $(\alpha, i')$ and $(\beta, j)$ -- with $\vzbar\link{\alpha}{i'}{\beta}{j'}$ and $\calE_{I,\beta}$ or with $\calE_{I,\alpha}$ and $\vzbar\subs$ -- result in positions that are within a distance $<\tau_\text{tol}$.

RelPose still constitutes a significant amount of data, so we introduce a parameter $\tau_\text{mdg}$ (minimum distance geometric verifications) which allows to throttle geometric verifications.
Geometric verification are skipped for matches which are close to already established relative poses.
$d^{\alpha \rightarrow \beta}$ is only sent for geometric verification to $\beta$ if there is no other $(\alpha, i')$ matched to a frame of $\beta$ such that $\ltnorm{\vt\of{\alpha}{i} - \vt\of{\alpha}{i'}} < \tau_\text{mdg}$.
This assumes that the VO and RelPose are consistent enough such that the resulting subset of retrievable $\mathcal{E}_S$ suffices to produce a globally-consistent state estimate.

\subsection{Decentralized Optimization}\label{dopt}
In our decentralized optimization, each robot estimates
its own trajectory using the available  measurements, and 
leveraging occasional
communication with other robots.
The pose of frame $(\alpha, i)$ is denoted with $\vxx\of{\alpha}{i}$.
 The trajectory of robot $\alpha$ is then denoted as $\vxx\of{\alpha}{} = [\vxx\of{\alpha}{1},\vxx\of{\alpha}{2}, \ldots]$. , where $\vxx\of{\alpha}{i} = (\MR\of{\alpha}{i}, \vt\of{\alpha}{i}) \in \SEthree$,  $\MR\of{\alpha}{i} \in \SOthree$ represents the 3D rotation, and $\vt\of{\alpha}{i} \in \Real{3}$ represents the 3D translation.

While our classification of the measurements (inter \vs ~ intra) is based on the robots involved in the 
measurement process, all relative measurements can be framed within the same measurement model.
Since all measurements correspond to noisy observations of the relative pose between a pair of poses, say 
 $\vxx\of{\alpha}{i}$ and $\vxx\of{\beta}{j}$, a general measurement model is:
\beq
\label{eq:measurementModel}
\begin{array}{c}
\vzbar\subs \doteq (\MRbar\subs, \vtbar\subs), \quad \text{with:} 
\left\{
\begin{array}{l}
\MRbar\subs = (\MR\of{\alpha}{i})\tran \MR\of{\beta}{j} \MR_\epsilon    \\
\vtbar\subs = (\MR\of{\alpha}{i})\tran ( \vt\of{\beta}{j} \!-\! \vt\of{\alpha}{i}  ) \!+\! \vt_{\epsilon}
\end{array} 
\right.                      
\end{array}
\eeq
where the relative pose measurement $\vzbar\subs$ includes the relative rotation measurements
$\MRbar\subs$, which describes the attitude $\MR\of{\beta}{j}$ with respect to the 
  reference frame $(\alpha,i)$,
``plus'' a random rotation $\MR_\epsilon$ (measurement noise), 
and the relative position measurement $\vtbar\subs$, which describes the position 
$\vt\of{\beta}{j}$ in the reference frame $(\alpha,i)$, plus random
 noise $\vt_{\epsilon}$. 

Assuming that the translation noise is distributed according to a zero-mean Gaussian with information matrix
 $\omegat \eye_3$, while the rotation noise follows a Von-Mises distribution with 
concentration parameter $\omegaR$, it is possible to demonstrate~\cite{Carlone15iros-duality3D}
that the ML estimate $\vxx \doteq \{(\MR\of{\alpha}{i},\vt\of{\alpha}{i}), \forall \alpha \in \Omega, \forall i\}$
can be computed as solution of the following optimization problem:
\bea
\label{eq:MRPGO}
\!\!\!\!
\min_{
\substack{
\vt\of{\alpha}{i} \in \Real{3}, 
\MR\of{\alpha}{i} \in \SOthree \\
\forall \alpha \in \robots, \forall i
}} 
 \sum_{(\alpha_i,\beta_j) \in \calE}  & 
 \omegat \normsq{ \vt\of{\beta}{j} \!\!-\! \vt\of{\alpha}{i} \!\!-\! \MR\of{\alpha}{i} \vtbar\subs }{}  
 \!\!+ \nonumber
 \\ 
 &  \frac{\omegaR}{2} \normsq{ \MR\of{\beta}{j} \!\!-\! \MR\of{\alpha}{i} \MRbar\subs}{\frob}  
\eea

As proposed in~\cite{choudhary2017distributed}, 
 we use a two-stage approach to solve the optimization problem in a distributed manner: we first solve a relaxed version of~\eqref{eq:MRPGO} and get an estimate for the rotations 
 $\MR\of{\alpha}{i}$ of all robots, and then we recover the full poses and top-off the result with a Gauss-Newton (GN) iteration.

In both stages, we need to solve a linear system where 
 the unknown vector can be partitioned into subvectors, such that each subvector contains the 
 variables associated to a single robot in the team. For instance, we can partition the vector $\vr$ in the first stage, 
  as $\vr = [\vr_\alpha, \vr_\beta, \ldots]$, such that $\vr_\alpha$ describes the rotations of robot $\alpha$.
  Similarly, we can partition  $\vp = [\vp_\alpha, \vp_\beta, \ldots]$ ($\vp$ represents the linearized poses) in the second stage, such 
  that $\vp_\alpha$ describes the trajectory of robot $\alpha$.
  Therefore, the linear systems in the first two stages can be framed in the general form $\MH \vy = \vg$:
\beq
\label{eq:partition}
\left[
\begin{array}{ccc}
\MH_{\alpha\alpha} & \MH_{\alpha\beta} & \ldots \\
\MH_{\beta\alpha} & \MH_{\beta\beta} & \ldots \\
\vdots & \vdots & \ddots
\end{array}
\right]
\left[
\begin{array}{c}
\vy_\alpha \\ \vy_\beta \\ \vdots 
\end{array}
\right]
=
\left[
\begin{array}{c}
\vg_\alpha \\ \vg_\beta \\ \vdots 
\end{array}
\right]
\eeq
where we want to compute the vector $\vy = [\vy_\alpha, \vy_\beta, \ldots]$ given 
the (known) block matrix $\MH$ and the (known) block vector $\vg$, partitioned according to the block-structure of $\vy$.

\newcommand{\indTwo}{\delta}

The linear system~\eqref{eq:partition} can be rewritten as:
 \[
\sum_{\indTwo \in \MOmega } \MH_{\alpha\indTwo} \vy_\indTwo = \vg_\alpha \qquad \forall \alpha \in \MOmega
 \] 
Taking the contribution of 
$\vy_\alpha$ out of the sum, we get:
\beq
\label{eq:distPreamble}
\MH_{\alpha\alpha} \vy_\alpha = - \!\!\!\sum_{\indTwo \in \MOmega \setminus \{\alpha\} } \!\!\! \MH_{\alpha\indTwo} 
\vy_\indTwo + \vg_\alpha
\qquad \forall \alpha \in \MOmega
\eeq
The set of equations~\eqref{eq:distPreamble} is the same as the original system~\eqref{eq:partition}, 
but clearly exposes the contribution of the variables associated to each robot.

The distributed Gauss-Seidel algorithm~\cite{Bertsekas89book} starts at an arbitrary initial estimate
$\vy\at{0} = [\vy_\alpha\at{0}, \vy_\beta\at{0}, \ldots]$ and
applies the following update rule, for each robot $\alpha \in \MOmega$:
\begin{multline}
\label{eq:sorIterations}
\!\!\! \!\!\!\vy\at{k+1}_\alpha = \!\!\MH_{\alpha\alpha}\inv \left(
  - \sum_{\indTwo \in \MOmega^+_\alpha } 
\!\!\! \MH_{\alpha\indTwo} \vy\at{k+1}_\indTwo
\!\! - \sum_{\indTwo \in \MOmega^-_\alpha } 
\!\!\! \MH_{\alpha\indTwo} \vy\at{k}_\indTwo  
 + \vg_\alpha \right)
\end{multline}
where $\MOmega^+_\alpha$ is the set of robots that already computed the $(k+1)$-th estimate, while
$\MOmega^-_\alpha$ is the set of robots that still have to perform the update~\eqref{eq:sorIterations},
excluding node $\alpha$ (intuitively: each robot uses the latest estimate).
We can see that if the sequence produced by the iterations~\eqref{eq:sorIterations} converges to a fixed point,
then such point  satisfies~\eqref{eq:distPreamble}, and indeed solves the original linear
system~\eqref{eq:partition}. 

\subsection{Making Optimization Work Continuously}

Since the optimization consists of two stages, and the map estimate at the end of the first stage and between iterations is not a consistent map state, the map estimate handled by DOpt cannot be considered as best estimate at any given time.
Furthermore, the optimization is not laid out to incorporate new measurements between iterations.
As a consequnce, the map estimate is concurrently modified by the VO and DOpt, which would lead to inconsistencies if unchecked.
This can be solved using optimistic concurrency control (OCC).
In OCC, each of the processes that concurrently operate on some data operate on their own copy.
Once a process is done, its copy is merged back to the reference state.
OCC is a central concept in version control tools such as SVN and GIT, and has recently been applied to distributed robotic systems in \cite{cieslewski2015map} and \cite{GaddIROS2016}.
In our method, the handling of VO and place recognition is considerd one process and decentralized optimization another.
We let VO and place recognition operate directly on the reference state.
Decentralized optimization, as described in \refsec{dopt} is executed episodically.
At the beginning of each episode, the robots consent on a reference time $t_e$.
During the episode, only the poses $\vxx\of{\alpha}{i}$ which satisfy $t\of{\alpha}{i} < t_e$ are optimized.
Once the optimization has finished, the $\vxx\of{\alpha}{i}$ of the reference state are replaced with the newly optimized estimates $\vxx\of{\alpha}{i}'$.
The remaining $\vxx\of{\alpha}{j}$ for which $t\of{\alpha}{j} > t_e$ are corrected with:
\begin{equation}
\vxx\of{\alpha}{j} \gets \MR \cdot \vxx\of{\alpha}{j} + \vt
\end{equation}
which satisfies
\begin{equation}
\vxx\of{\alpha}{e}' = \MR \cdot \vxx\of{\alpha}{e} + \vt
\end{equation}
with $(\alpha, e)$ referring to the frame recorded just before $t_e$.

\section{Experiments}
\label{sec:experiments}

In the experiments, we seek to find out how much data the individual components and the overall system use and how accurate the state estimate is as time progresses.
This will give us an idea of the applicability and scalability of the system, and will let us identify which components are most worthwhile to be optimized in the future.

\subsection{Data Exchange Evaluation}
The components of our system that exchange data are decentralized visual place recognition, relative pose estimation and decentralized optimization.
Each of these components could potentially be further developed to use less data.
Thus, we record data transfer for each component individually.
Data transfers for the individual components are:
\begin{align}
\bigcup_\text{episode} d_\text{DOpt}^{\alpha \rightarrow \beta} &= \{\bigcup_\text{rot. iters.}\MR\subs, \bigcup_\text{pose iters.}\vzbar\subs\} \\
d_\text{DVPR}^{\alpha \rightarrow \delta} &= (\alpha, i, \vv\of{\alpha}{i}),\quad d_\text{DVPR}^{\delta \rightarrow \alpha} = (\beta, j) \\
d_\text{RelPose}^{\beta \rightarrow \alpha} &= (\vzbar\subs, \vxx\of{\beta}{j'}^{-1}\vxx\of{\beta}{j})
\end{align}
with $d_\text{RelPose}^{\alpha \rightarrow \beta}$ from \eqref{exchdata} and where $\vxx\of{\beta}{j'}^{-1}\vxx\of{\beta}{j}$ is sent for the relative pose consistency check.
The size of data is calculated accordingly, using $6\cdot 8$ bytes for poses $\vzbar\subs$, $9 \cdot 8$ bytes for rotations $\MR\subs$, $8\cdot D_\text{NetVLAD}$ bytes for NetVLAD vectors $\vv\of{\alpha}{i}$, $1$ byte for robot indices, $4$ bytes for frame indices, $2$ bytes for visual word indices $w_k$ and $3 \cdot 4$ bytes for landmarks $\vp_k$.

\subsection{Accuracy Evaluation}

As it is typically done for SLAM systems, we evaluate the accuracy of our system.
Unlike of other SLAM systems, however, we evaluate the evolution of accuracy over time.
This allows us to better characterize the system.
Accuracy is measured using the average trajectory error (ATE).
The ATE requires alignment to the ground truth trajectory, which is only meaningful for connected components.
Thus, we report the ATE for different connected components individually.

\subsection{Parameter Studies}
A parameter that we find of particular interest is $\tau_\text{mdg}$, which determines the distance between an established relative pose $\vzbar\subs$ and the next frame of $\alpha$ to be sent to $\beta$ for geometrical verification.
$\tau_\text{mdg}$ can significantly reduce the data to be transmitted for RelPose and DOpt, but it is important to verify that this does not happen at the cost of accuracy.
We vary $\tau_\text{mdg}$ and see how it affects data transmission and ATE.

Another interesting parameter is the used NetVLAD dimension $D_\text{NetVLAD}$.
Since the last layer of NetVLAD performs principal component analysis, $D_\text{NetVLAD}$ can be tuned arbitrarily up to $4096$ dimensions.
The lower $D_\text{NetVLAD}$, the lower the DVPR traffic, but also the lower its precision.

\section{Results}
\label{sec:results}

\begin{table}
\centering
\begin{tabular}{|l|l|l|l|l|l|}
\hline
& {\bf Sec.} & {\bf value} & & {\bf Sec.} & {\bf value} \\
\hline
$D_\text{NetVLAD}$ & \ref{sec:dvpr} & $128$ & $\tau_\text{cdist}$ & \ref{sec:relpose} & $20m$ \\
\hline
$\tau_\text{NetVLAD}$ & \ref{sec:dvpr} & $0.1$ & $\tau_\text{tol}$ & \ref{sec:relpose} & $4m$ \\
\hline
$\tau_\text{inliers}$ & \ref{sec:relpose} & $20$ & $\tau_\text{mdg}$ & \ref{sec:relpose} & $0m$ \\
\hline
$\tau_\text{loss}$ & \ref{sec:relpose} & $3m$ & & & \\
\hline
\end{tabular}
\vspace{2mm}
\caption{Default parameter values}
\label{tab:params}
\vspace{-8mm}
\end{table}
We evaluate our system on the KITTI dataset \cite{geiger2013vision}.
Like in \cite{cieslewski2017efficient}, we split the dataset into $|\MOmega|$ parts to simulate trajectories recorded by different robots.
\begin{figure}
  \centering
    \includegraphics[width=.8\columnwidth]{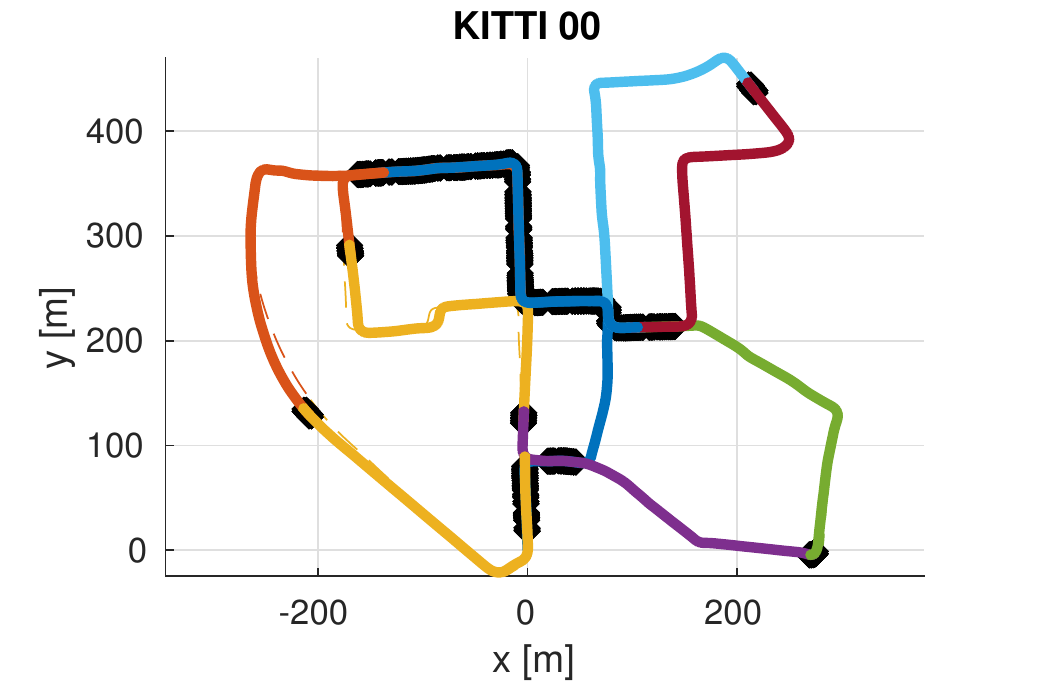}
    \vspace{-4mm}
    \caption{Ten sub-trajectories of KITTI 00 after running our method with the parameters in \reftab{tab:params}.
  Each color represents an individual robots trajectory, place matches are marked with bold black dots.
  The aligned ground truth is indicated with dashed lines.}
  \label{kitti}
  \vspace{-4mm}
\end{figure}    
\reffig{kitti} shows the used sub-trajectories of KITTI 00 and the end result of running our system with the parameter values shown in \reftab{tab:params}.
\begin{figure}   
  \centering
    \includegraphics[width=.8\columnwidth]{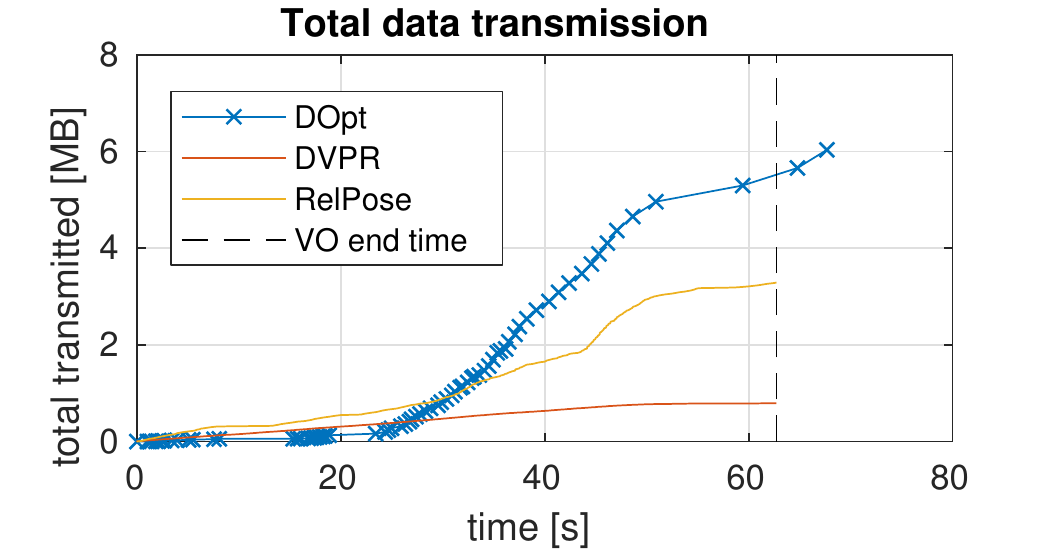}
    \vspace{-4mm}
    \caption{Data transmission over time for the three system components:
  decentralized optimization (DOpt), decentralized visual place recognition (DVPR) and relative pose estimation (RelPose), using the parameters in \reftab{tab:params}.
  DOpt continues to run after visual odometry (VO) has ended until it has incorporated all data produced by VO.}
  \label{data-time}
  \vspace{-4mm}
\end{figure}
The corresponding data transmission is shown in \reffig{data-time}.
Note that we plot the cumulative data transmission up to a given time.
This directly exhibits the full amount of transmitted data and makes it easy to see time-varying bandwidth (slope).
DOpt and RelPose require the most data transmission, but as we will see, DOpt traffic can be significantly reduced, while RelPose traffic remains high.

\begin{figure}
  \centering
  \vspace{-4mm}
  \includegraphics[width=\columnwidth]{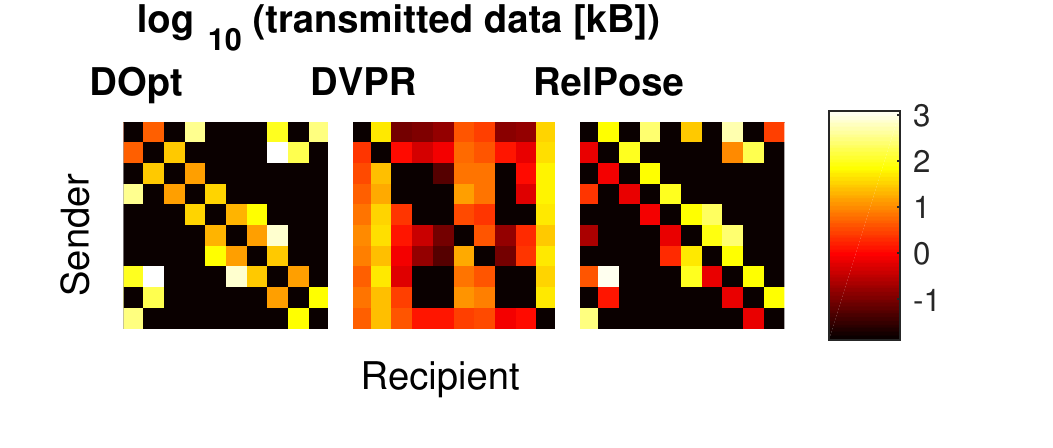}
  \vspace{-10mm}
  \caption{Data exchanged between individual pairs of robots.
  Row numbers indicate sender and column numbers receiver.
  DOpt, DVPR and RelPose traffic is visualized separately and separated by green lines.}
  \label{data-robots}
\end{figure}
\reffig{data-robots} shows the total data transmission between pairs of robots.
This data transmission is highly uneven and could require consideration in multi-hop networks, but that is outside of the scope of this paper.
DOpt and RelPose highly depend on the trajectory overlap between robots.
The higher the overlap, the more data is exchanged by these components.
Decentralized visual place recognition traffic is also uneven, but in a different way.
It is due to the clustering that assigns NetVLAD cluster centers to robots.
What we see in \reffig{data-robots} is that robots 2, 6 and 7 are assigned the $\vv$ that represent the majority of NetVLAD descriptors in KITTI 00.
This unevenness can be mitigated either by randomly assigning several clusters per robot or by training the clustering on a dataset that is very similar to the data at hand, in terms of the scope of visual appearances of keypoints.
A detailed study of this problem is provided in \cite{cieslewski2017efficient2}.

\begin{figure}
  \centering
  \includegraphics[width=.8\columnwidth]{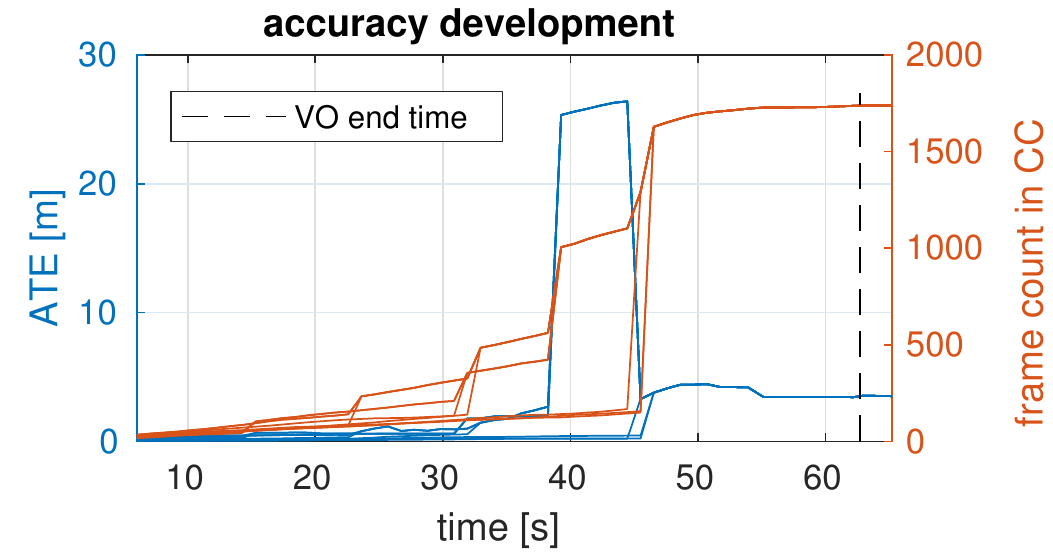}
  \caption{Average trajectory error (ATE) and size of the connected component (CC) each of the ten robots is in.
  As the CCs of two robots converge, so does the line representing the frame count and ATE within these CC.
  Note that accuracy does not change significantly in the optimizations executed after visual odometry (VO) has ended -- the state estimate reaches its final accuracy already before all the data is considered for optimization.}
  \label{accuracy}
  \vspace{-4mm}
\end{figure}
\reffig{accuracy} illustrates how the trajectories of the different robots fuse as time progresses, as well as the accuracy of the resulting connected components.
Seven merge events of connected components are apparent around seconds 15, 23, 32, 33, 39, 46 and 47, the remaining two merges occur at the beginning of the experiment and are not visible in \reffig{accuracy}.
At each merge event, there is a jump in connected component size and the corresponding lines in the plot get fused.
In early merge events, the ATE grows as the drift of individual components is compounded with noise in relative pose measurements.
This is most pronounced at the merge event around second 39.
In later merge events, and as more relative measurements are added, loops are closed and the ATE decreases again, until it stabilizes around a value of 4 meters.
%
%
This is well below 1\% of the overall trajectory length.

\begin{figure}
  \centering
  \includegraphics[width=.8\columnwidth]{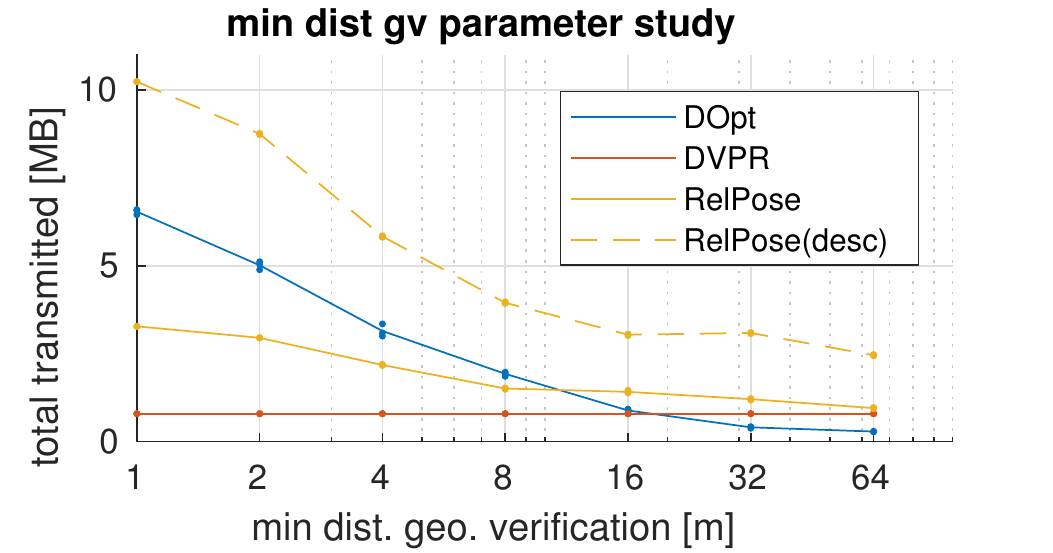}
  \caption{Effect of changing $\tau_\text{mdg}$, the parameter that controls the distance between an established relative pose $\vzbar\subs$ and the next frame of $\alpha$ to be sent to $\beta$ for geometrical verification.
  As we can see, $\tau_\text{mdg}$ can be used to significantly throttle the required bandwidth.
  We have verified that this happens without a significant effect on the ATE, which is around $4m$ throughout.
  The dashed line indicates RelPose data transmission if descriptors are used for keypoint association rather than visual word indices.}
  \label{mdgv}
  \vspace{-4mm}
\end{figure}
\reffig{mdgv} shows how increasing the minimum distance between geometric verifications $\tau_\text{mdg}$ can significantly reduce bandwidth requirements without negatively affecting accuracy.
\reffig{kitti-60} shows the place matches that remain and \reffig{data-time-60} data transmission over time with $\tau_\text{mdg} = 60m$.
%

\begin{figure}
  \centering
  \includegraphics[width=.8\columnwidth]{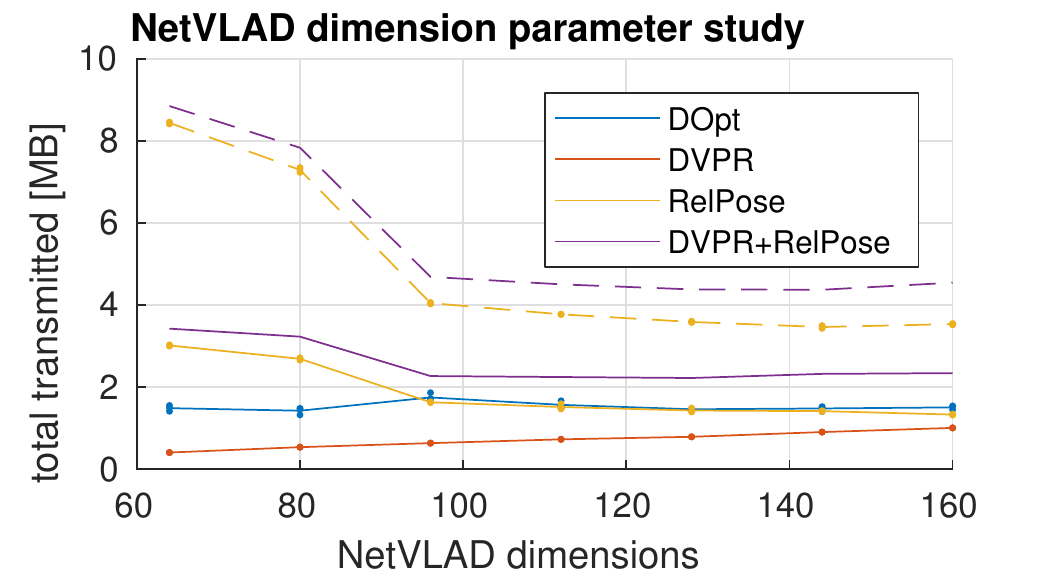}
  \caption{Data transmission and ATE as the NetVLAD dimensionality is varied, with $\tau_\text{mdg} = 10$.
  Dashed lines indicate RelPose data transmission if keypoints are matched with descriptors, rather than words.
  The ATE is around $4m$ throughout the samples.}
  \label{nvdim}
  \vspace{-4mm}
\end{figure}
\reffig{nvdim} shows how varying the NetVLAD dimension affects the size of transmitted data, with $\tau_\text{mdg} = 10m$.
Unsurprisingly, DVPR traffic is quasi-linear with respect to the NetVLAD dimension.
However, using a too low dimension for NetVLAD results in lower precision, which results in more failed relative pose estimates.
Since $\tau_\text{mdg}$ only suppresses relative pose queries if a relative pose query has previously been successful, this results in a significant increase of RelPose data transmission.
Conversely, increasing NetVLAD dimensions beyond a certain point does not increase precision significantly.
In \reffig{nvdim} we can see that an efficient trade-off is reached with $D_\text{NetVLAD} > 100$.

\begin{figure}
  \centering
  \includegraphics[width=.5\columnwidth]{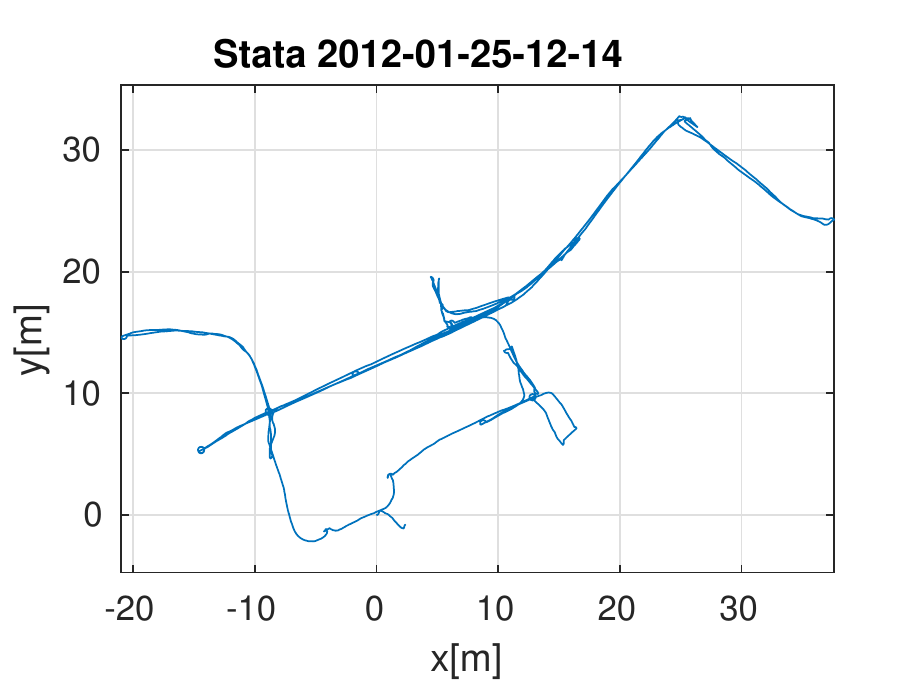}
 \vspace{-3mm}
  \caption{Groundtruth trajectory for one of the sequences in the MIT Stata Center Dataset \cite{Fallon13dataset}.}
  \label{stata_gt}
  \vspace{-4mm}
\end{figure}

\begin{figure}
  \centering
  \includegraphics[width=.48\columnwidth]{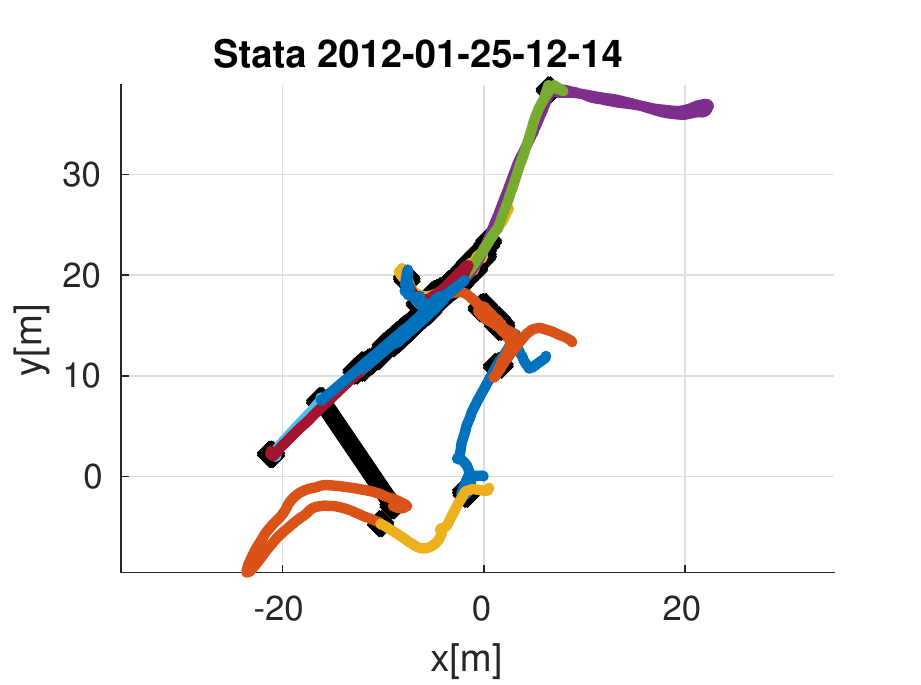}
  \includegraphics[width=.48\columnwidth]{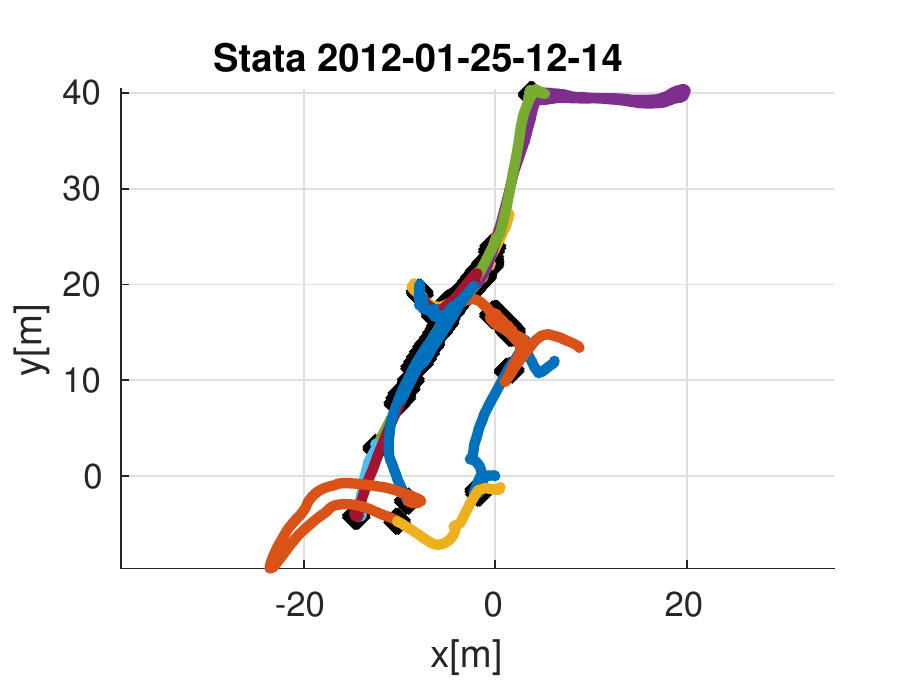}
   \vspace{-2mm}
  \caption{Ten sub-trajectories of the MIT Stata Center Dataset \emph{before} and \emph{after the final loop closure}, estimated using our method.
  Each color represents an individual robots trajectory, place matches are marked with bold black lines.}
  \label{stata_oops}
  \vspace{-4mm}
\end{figure}

We also evaluated our system on one of the sequences in MIT Stata Center Dataset \cite{Fallon13dataset}. 
\reffig{stata_gt} shows the groundtruth trajectory. 
Similar to the KITTI 00 dataset, we split the sequence into $|\MOmega|$ parts to simulate trajectories recorded by different robots.
\reffig{stata_oops} shows the sub-trajectories estimated using our method before and after the final loop closure.
As we can see, the final estimate given by our method is not visually similar to the groundtruth estimate (\reffig{stata_gt}). 
The main reason for this mismatch is that the graph structure in the case of MIT Stata Center dataset is closer to a chain topology. Due to the chain topology, decentralized optimization requires more iterations to reach consensus among robot trajectories on either ends of the chain and therefore finds it difficult to converge to the centralized estimate. Finding sub-trajectories which follow certain graph topologies (chain vs grid) and are favorable for decentralized optimization can be a useful future extension. 


\section{Conclusion}
\label{sec:conclusion}

We have presented a new integrated decentralized visual SLAM algorithm that is based on state-of-the-art components.
The system has been characterized using publically available data and we have explored how data transmission can be reduced to a minimum.
Based on our results, we believe that future developments of decentralized visual SLAM will focus on one hand on even more data-efficient data association and on the other hand on more robust decentralized optimization.

\bibliographystyle{IEEEtran} 
\bibliography{references,refs}

\addtolength{\textheight}{-3cm}   

\end{document}